\documentclass[a4paper]{article}

\usepackage{INTERSPEECH2018}

\usepackage{amssymb}
\usepackage{enumitem}
\usepackage{xcolor}
\usepackage{url}
\setlist[description]{leftmargin=0.0cm,labelindent=0.0cm}

\usepackage{cleveref}
\crefname{section}{\S}{\S\S}
\Crefname{section}{¤\S}{\S\S}
\newcommand{\ms}[1]{ }
\newcommand{\gn}[1]{ }

\title{Self-Attentional Acoustic Models}
\name{Matthias Sperber$^1$, Jan Niehues$^1$, Graham Neubig$^2$, Sebastian St\"uker$^1$, Alex Waibel$^{12}$}
\address{
  $^1$Karlsruhe Institute of Technology\\
  $^2$Carnegie Mellon University}
\email{\{first\}.\{last\}@kit.edu, gneubig@cs.cmu.edu}

\begin{document}

\maketitle
\begin{abstract}
Self-attention is a method of encoding sequences of vectors by relating these vectors to each-other based on pairwise similarities.
These models have recently shown promising results for modeling discrete sequences, but they are non-trivial to apply to acoustic modeling due to computational and modeling issues.
In this paper, we apply self-attention to acoustic modeling, proposing several improvements to mitigate these issues:
First, self-attention memory grows quadratically in the sequence length, which we address through a downsampling technique.
Second, we find that previous approaches to incorporate position information into the model are unsuitable and explore other representations and hybrid models to this end.
Third, to stress the importance of local context in the acoustic signal, we propose a Gaussian biasing approach that allows explicit control over the context range.
Experiments find that our model approaches a strong baseline based on LSTMs with network-in-network connections while being much faster to compute.
Besides speed, we find that interpretability is a strength of self-attentional acoustic models, and demonstrate that self-attention heads learn a linguistically plausible division of labor.%
\footnote{Code at \url{http://msperber.com/research/self-att}}
\end{abstract}
\noindent\textbf{Index Terms}: speech recognition, acoustic model, self-attention

\section{Introduction}
In order to transform an acoustic signal into a useful abstract representation, acoustic models must take into account the complex interplay of local and global dependencies in an acoustic signal. At a local, temporally constrained level, we observe concrete linguistic events (phonemes), while at a global level the signal is influenced by factors such as channel and voice properties. Traditional acoustic models reflect this intuition about global and local dependencies by first applying a normalization phase, a global operation that aims at producing invariance with respect to channel and speaker characteristics. After this, traditionally a hidden Markov model is applied over polyphones, modeling only local dependencies (beads-on-a-string view \cite{ostendorf1999moving}). This restriction has in part been motivated by the intuition that global effects should be removed from the signal at this stage.

However, the empirical success of recurrent neural networks (RNNs) for acoustic modeling \cite{sak2014long} has challenged this intuition and indicated that considering the global context is still beneficial at this stage. Unfortunately, RNNs suffer from slow computation speed and may not be able to optimally exploit long-range context. Self-attentional architectures \cite{cheng2016a,Parikh2016,Lin2017} have recently shown promising results as an alternative to RNNs for modeling discrete sequences \cite{Vaswani2017}. These models relate different positions in a sequence by computing pairwise similarities, in order to compute a higher level representation of the sequence. Self-attention is attractive (1) computationally because it can be efficiently implemented through batched tensor multiplication, and (2) from a modeling perspective because it allows direct conditioning on both short range-context and long-range context, without the need to pass information through many intermediate states as is the case with RNNs.

In this paper, we explore self-attentional architectures for acoustic modeling, by using the listen-attend-spell model \cite{Chan2016} and replacing its pyramidal encoder component with self-attention.\footnote{We also refer to independent work that has concurrently addressed similar questions \cite{Povey2018}.} In order to make self-attentional architectures work for acoustic modeling, several challenges must be addressed. First, self-attention computes the similarity of each pair of inputs, so the amount of memory grows quadratically with respect to the sequence length. This is problematic for modeling acoustic sequences, because these can get very long, e.g.\ our training utterances contain up to 2026 frames (800 on average). To address this issue, we apply downsampling by reshaping the sequence before self-attentional layers.

The second challenge is incorporating positional information into the model. Unlike an RNN, self-attention has no inherent mechanism of modeling sequence position. Vaswani et al.\ \cite{Vaswani2017} propose an additive trigonometric position encoding, which is problematic in the case of acoustic modeling because our inputs are fixed speech features rather than flexibly learned word embeddings. While concatenating positional embeddings instead provides some remedy, we find it necessary to design a hybrid self-attention/RNN architecture to obtain good results.

The third challenge is effective modeling of context relevance. Speech frames contain much less information than words and it is therefore more difficult to estimate the importance of pairs of frames with respect to each other. Based on the intuition that locality of context plays a special role in acoustic modeling, we propose to apply diagonal Gaussian masks with learnable variance to attention heads. This gives attention heads more control over context relevance and improves word error rates consistently, by up to 1.59\%. We observe that while bottom layer attention heads converge toward diversity in context range, higher layers use long-range context.

Attention mechanisms improve the often criticized poor interpretability of neural end-to-end models because they enforce an explicit expression of dependencies. Self-attention brings this interpretability inside the encoder, making it possible to examine how speech is encoded before making the final recognition decisions. Our analysis reveals that different attention heads measure similarity along different linguistically plausible dimensions such as phoneme clusters, indicating that they function in part to reduce acoustic variability by establishing averaged versions of matching acoustic events across the utterance.

\section{Attentional Models for ASR}

\subsection{Listen, Attend, Spell}
\label{sec:las}
Our ASR model is based on the listen-attend-spell model \cite{Chan2016}, an attentional encoder-decoder model \cite{Kalchbrenner2013,Sutskever2014,Bahdanau2014} where the encoder component serves as an acoustic model, taking speech features as input. Because acoustic sequences are very long, the encoder performs downsampling to make memory and runtime manageable. This is achieved through a pyramidal LSTM (Fig.~\ref{fig:architectures}a), a stack of LSTM layers where pairs of consecutive outputs of a layer are concatenated before being fed to the next layer, such that the number of states is halved between layers.

\subsection{Existing Encoders for Speech}
\label{sec:nin}
\begin{figure}[t]
  \centering
  \includegraphics[width=0.95\linewidth]{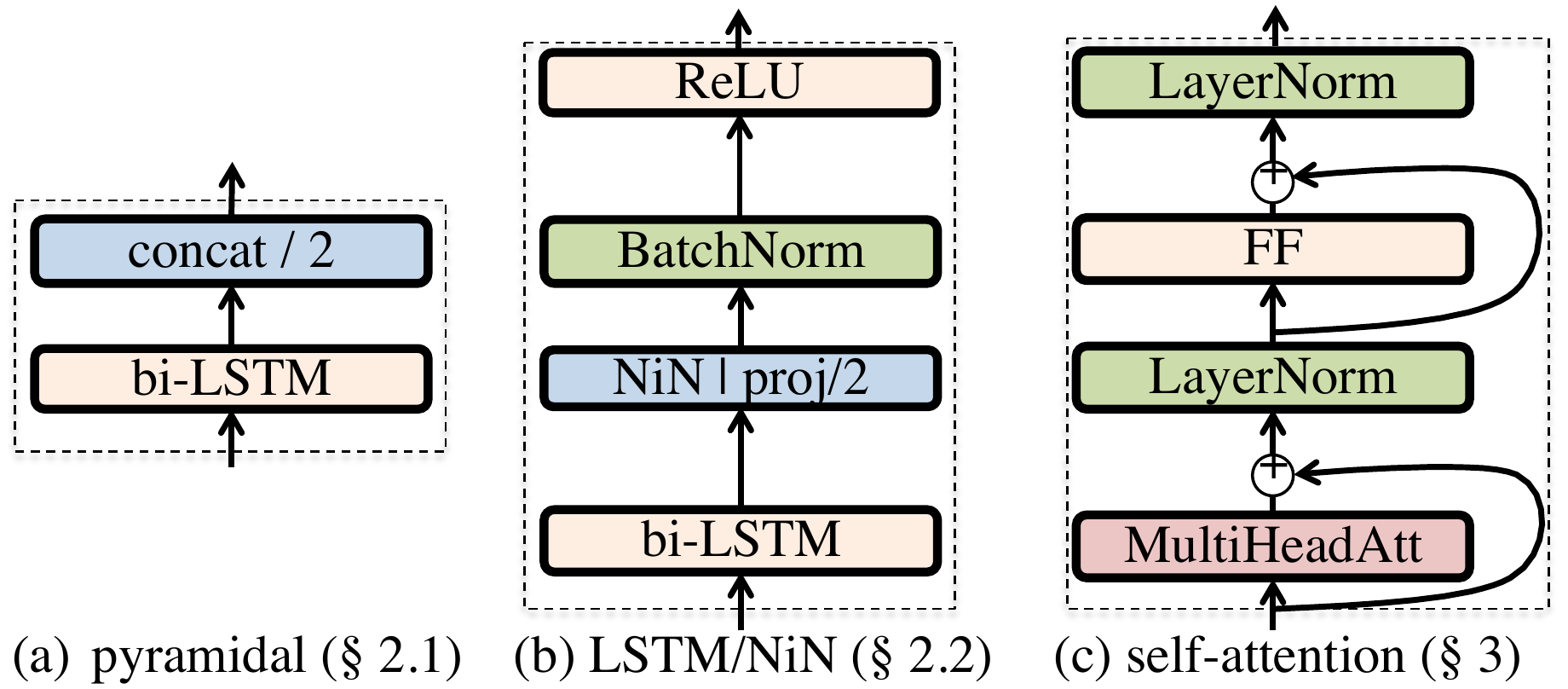}
  \caption{Block diagrams of baselines and the core model.}
  \label{fig:architectures}
\end{figure}

Several improvements over the pyramidal LSTM encoder have been proposed \cite{Zhang2017e}. As a second baseline, we employ a state-of-the-art model \cite{Zhang2017e} that stacks blocks consisting of an LSTM, a network-in-network (NiN) projection, and batch normalization (Fig.~\ref{fig:architectures}b). The top LSTM/NiN block is extended by a final LSTM layer. NiN denotes a simple linear projection applied at every time step, possibly performing downsampling by concatenating pairs of adjacent projection inputs.

\section{Self-Attentional Acoustic Models}
\label{sec:sa}

Self-attention is applied to a sequence of state vectors and transforms each state into a weighted average over all the states in the sequence, with more relevant states being given more influence. The underlying intuition is that states at each time step should be conditioned on the most relevant states across the whole sequence. Our basic form of self-attention follows Vaswani et al. \cite{Vaswani2017}, where relevance is measured by computing dot product similarity after applying a linear projection to both vectors. For acoustic sequences, neighboring frames are naturally similar if they represent parts of the same acoustic event. When an event with similar acoustic characteristics appears at different places in an utterance, those occurrences would be deemed relevant, as well. Following \cite{Vaswani2017} we use 8 attention heads where each head can compute this similarity independently.

Our model is specifically computed as follows (Fig.~\ref{fig:architectures}c):

\begin{align}
Q_{i}{=}XW_{i}^\text{Q}, \label{eq:qkv}
K_{i}{=}XW_{i}^\text{K},
V_{i}{=}XW_{i}^\text{V} \quad\quad\\
\text{head}_{i}=\mathrm{softmax}(\frac{Q_{i}K_{i}^\text{T}}{\sqrt{d}})V_{i} \quad\forall i \label{eq:head}\\
\text{MultiHeadAtt}=\mathrm{concat}(\text{head}_{1},\text{head}_{2},\ldots) \\
\text{MidLayer}=\mathrm{LayerNorm}\left[\text{MultiHeadAtt}+X\right] \\
\text{SAL}=\mathrm{LayerNorm}\left[\mathrm{FF}\left(\text{MidLayer}\right)+\text{MidLayer}\right] \label{eq:SA}
\end{align}
Here, $X{\in}{\mathbb{R}^{l\times d}},Q_i,K_i,V_i{\in}{\mathbb{R}^{l\times d/n}}$ denote inputs and their query/key/value transformations for attention heads indexed by $i{\in}\{1,\cdots,8\}$, sequence length $l$, and hidden dimension $d$. $\text{SAL}$ denotes the final output of the self attention layer. $W_{i}^\text{Q}, W_{i}^\text{K}, W_{i}^\text{V}\in\mathbb{R}^{d\times d/n}$ are parameter matrices. $\text{FF}$ is a position-wise feed-forward network intended to introduce additional depth and nonlinearities, defined as $\text{FF}(x){=}\max\left(0,xW_1+b_1\right)W_2+b_2$. $\text{LayerNorm}$ is according to \cite{Ba2016a}.

\section{Tailoring Self-Attention to Speech}

\subsection{Downsampling}

To introduce downsampling so that the model described in \cref{sec:sa} fits in memory, we apply a reshaping operation before every self-attention block. This reduces the sequence length by a factor $a$ and increases the vector state dimension accordingly:

$$X\in \mathbb{R}^{l\times d}\rightarrow_\text{reshape}\hat{X}\in \mathbb{R}^{\frac{l}{a}\times ad}$$

We then compute (\ref{eq:qkv}) through (\ref{eq:SA}) as before, with the shape of weight matrices adjusted to $W_{i}^\text{Q}, W_{i}^\text{K}, W_{i}^\text{V}\in\mathbb{R}^{da\times d/n}$. This reduces the memory consumption of the attention matrix by factor $a^2$. It is crucial to apply reshapes also before the bottom layer so that the large bottom attention matrix is scaled down. Note that this approach is very similar to downsampling as in the pyramidal LSTM, except that it is applied to a sequence feature matrix instead of per-timestep, and also applied before the bottom layer.





\subsection{Position Modeling}
\label{sec:pos_modeling}

Position information is crucial in sequence-to-sequence models, but self-attention is completely agnostic to sequence positions. Prior works added trigonometric position encodings \cite{Vaswani2017} or learned position embeddings \cite{Gehring2017} to input vectors, but we found that this approach does not work well for acoustic sequences. This is intuitive, as the inputs are fixed feature vectors rather than trainable word embeddings, making it difficult for the model to separate position and content for each state. 

\subsubsection{Concatenated Position Representation}
A straight-forward solution to enable separation of position and content for fixed inputs is to concatenate position representation instead of using a sum. We explore three variants: First, concatenating trigonometric encodings \cite{Vaswani2017} to the input feature vectors. Second, concatenating learned embeddings \cite{Gehring2017} to inputs. Third, concatenating separately learned position embeddings to the queries and keys ($Q$,$K$ in Equation \ref{eq:qkv}) so that the key and query position can be taken into account when computing relevance at each layer. 


\subsubsection{Hybrid Models}
RNNs are effective at keeping track of positional information. We can exploit this by introducing recurrent layers into our encoder. We explore two alternatives:

{\bf Stacked hybrid model}. Here, we stack 2 LSTM/NiN blocks (Fig.~\ref{fig:architectures}b) without downsampling, followed by a final LSTM, on top of our self-attention layers. This approach does not make the self-attention layers themselves position-aware, but the final encoder states are position-aware. Reversing the order of self-attention and LSTM/NiN is also conceivable but would compromise speed because slow recurrent computations are applied before downsampling.

{\bf Interleaved hybrid model}. Another option is to replace the feed-forward operation ($\text{FF}$ in Equation \ref{eq:SA}) by an LSTM. Note that this introduces LSTMs before the sequence is fully downsampled and therefore compromises some of the speed gains. On the other hand, it allows the higher self-attention layers to take advantage of position information encoded by lower interleaved LSTMs.

\subsection{Attention Biasing}
\label{sec:att_bias}

Self-attention allows direct conditioning on the whole sequence, but it is unclear to what extent this is beneficial for our acoustic model. While context required to model polyphones may span only a relatively small temporal window, remaining channel and speaker properties may require long-range context. To account for the special role of context locality in acoustic modeling, we introduce an explicit way of controlling the context range by using a bias matrix $M\in \mathbb{R}^{l\times l}$ and computing $\text{head}_{i}=\mathrm{softmax}(\frac{Q_{i}K_{i}^{T}}{\sqrt{d}}+M)V_{i}$. By setting values around the diagonal of this mask to a higher value, we can bias the self-attention toward attending in a local range around each frame.

\subsubsection{Local Masking}

We can apply hard masking by setting $M$ as an inversely banded matrix of bandwith $b\in \mathbb{N}_\text{odd}$ with 
$$M_{jk}=\begin{cases}0 & |j-k|<\frac{b}{2}\\\text{-\ensuremath{\infty}} & \text{else} \end{cases}.$$

As a result, all attention weights outside the band are set to 0, so that the self-attention is restricted to a local region of size $b$. The hyperparameter $b$ can be set prior to training such that the model effectively attends to a range similar to polyphone context in hidden Markov models. Notice the similar idea explored concurrently in independent work \cite{Povey2018}.

\subsubsection{Gaussian Bias}

For more flexibility, we use a soft Gaussian mask by defining 
$$M_{jk}=\frac{-(j-k)^{2}}{2\sigma^{2}}.$$
$\sigma$ is a trainable standard deviation parameter. It is learned separately for each attention head so that the context range can differ between attention heads. Besides more modeling expressiveness, the learned variances can also be inspected and may help us to understand and interpret the model.\footnote{To overcome trainability issues and encourage the optimizer to adjust the variance parameter, we found it necessary to re-parametrize it using $\tau^{2}=\sigma$ and optimize $\tau$ via back-propagation.} Note that this bears some resemblance to prior work \cite{Im2017} who use a {\it linear} distance map instead of a Gaussian and do not include trainable parameters, making their model less flexible and less interpretable.

\section{Experimental Setup}

We focus our experiments on the TEDLIUM corpus \cite{Rousseau2014}, a widely used corpus of 200h of recorded TED talks, with the development split used as validation data. Our implementation is based on the XNMT toolkit, with which we have previously demonstrated competitive ASR results on two benchmarks \cite{neubig18xnmt}.

The training settings follow \cite{neubig18xnmt} where relevant. 
We extract 40-dimensional Mel filterbank features with per-speaker mean and variance normalization using Kaldi \cite{povey2011kaldi}. We exclude utterances longer than 1500 frames to keep memory requirements manageable. The encoder-decoder attention is MLP-based, and the decoder uses a single LSTM layer. The number of hidden units is 128 for the encoder-decoder attention MLP, 64 for target character embeddings, and 512 elsewhere unless otherwise noted. The model uses input feeding \cite{Luong2015}, variational recurrent dropout with probability 0.2 and target character dropout with probability 0.1 \cite{Gal2016}. We apply label smoothing \cite{Szegedy2016} and fix the target embedding norm to 1 \cite{Nguyen2017}. For inference, we use a beam size of 20 and length normalization with exponent 1.5. Self-attention layers use a hidden dimension of 256 and feed-forward dimension of 256, and attention dropout with probability 0.2. When LSTMs are part of the encoder, we use bidirectional LSTMs with 256 hidden units per direction. Concatenated position representation vectors are of size 40.

The vocabulary consists of the 26 English characters, apostrophe, whitespace, and special start-of-sequence and unknown-character tokens. We set the batch size dynamically depending on the input sequence size such that the average batch size was 24 (18 for LSTM-free models). We use Adam \cite{Kingma2014} with initial learning rate of 0.0003, decayed by 0.5 when validation WER did not improve over 10 epochs initially and 5 epochs after the first decay.

\section{Quantitative Results}

\subsection{Comparison to Baselines}
The first set of experiments compares the proposed hybrid models to the baselines. The results are summarized in Table~\ref{tab:baselines}. We observe similar word error rates, with the interleaved model outperforming the stacked model and outperforming the pyramidal LSTM baseline on the development data but not the test data. The LSTM/NiN baseline was strongest. In terms of training speed, the stacked model is fastest by a large margin, followed by the interleaved model and the LSTM/NiN model.
To confirm that the attention mechanism is actually contributing to the hybrid model and not just passing on activations, we performed a sanity check by training a stacked hybrid model with attention scores off the diagonal set to $-\infty$, and observed a drop of 1.25\% absolute WER.

\begin{table}[t]
  \centering
  \caption{Comparison to baselines. Training speed  (char/sec) was measured on a GTX 1080 Ti GPU.}
\begin{tabular}{|c|c|c|c|c|}
\hline 
model & dev WER & test WER & char/sec \tabularnewline 
\hline 
\hline 
pyramidal & 15.83 & 16.16 & 1.1k \tabularnewline 
LSTM/NiN & 14.57 & 14.70 & 1.1k \tabularnewline 
\hline 
stacked hybrid & 16.38 & 17.48 & 2.4k \tabularnewline 
interleaved hybrid & 15.29 & 16.71 & 1.5k \tabularnewline 
\hline 
\end{tabular}
\label{tab:baselines}
\end{table}

\subsection{Position Modeling}

Next, we evaluate the different approaches to position modeling (\cref{sec:pos_modeling}). The results are summarized in Table~\ref{tab:pos_mod}. When using additive positional encodings the model diverged, while concatenating embeddings converged, albeit to rather poor optima. The key/query positional embeddings in isolation diverged, and combination with concatenated input embeddings did not improve results. Only the hybrid models were able to obtain results comparable to the baselines. We also tried combining hybrid models with positional embeddings, but did not see improvements over the model without positional embeddings.


\begin{table}[t]
  \centering
  \caption{WER results on position modeling.}
\begin{tabular}{|c|c|c|c|c|}
\hline 
model & dev & test \tabularnewline 
\hline 
\hline 
add (trig.) & \multicolumn{2}{c|}{diverged}\tabularnewline
\hline 
concat (trig.) & 30.27 & 38.60  \tabularnewline  
\hline 
concat (emb.) & 29.81 & 31.74 \tabularnewline 
\hline 
stacked hybrid & 16.38 & 17.48 \tabularnewline 
\hline 
interleaved hybrid & 15.29 & 16.71 \tabularnewline 
\hline 
\end{tabular}
\label{tab:pos_mod}
\end{table}

\subsection{Attention Biasing}

This set of experiments tests the effect of introducing explicit attention biases that enable the model to control its context range (\cref{sec:att_bias}). The local diagonal mask was set to constrain the context to a window of 5 time steps, and Gaussian biasing variances were initialized to 9 (small setting) or 100 (large setting). Results are summarized in Table~\ref{tab:biased_att}. For the stacked model, it can be seen that the biasing helps in general. The strongest model variant was the learnable Gaussian mask. Interestingly, it was important to initialize the Gaussian to possess a large variance. We hypothesize that this improves gradient flow early on in the model training, similar to how initializing LSTM forget gate biases to 1 (no forgetting) improves results \cite{Jozefowicz2015}. The interleaved hybrid model shows similar trends. Note that the sometimes inconsistent ordering between dev and test results can be explained by the fact that the TEDLIUM dev set is relatively small with only 500 utterances.

\begin{figure}[t]
  \centering
  \includegraphics[width=0.45\linewidth]{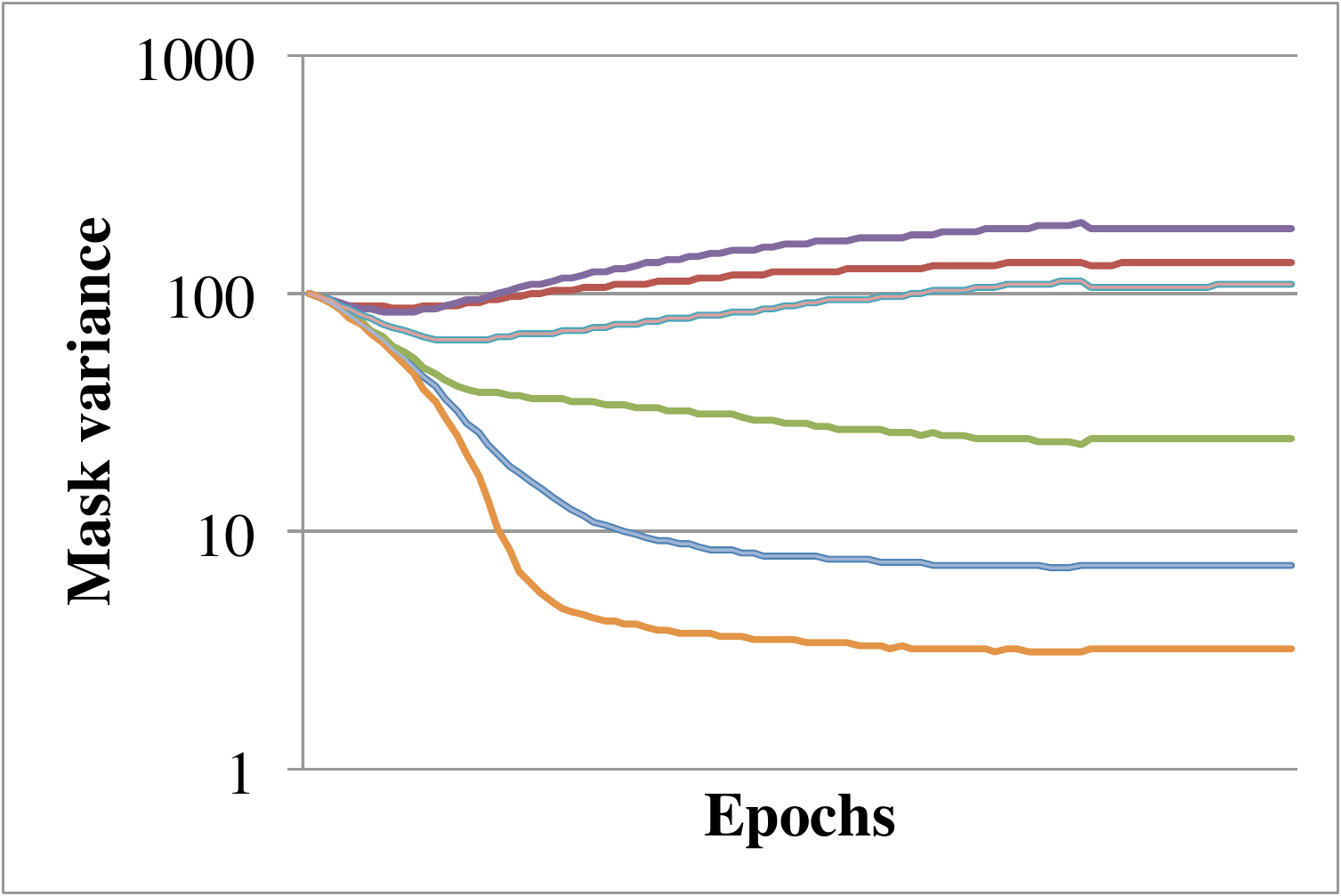}
  \includegraphics[width=0.45\linewidth]{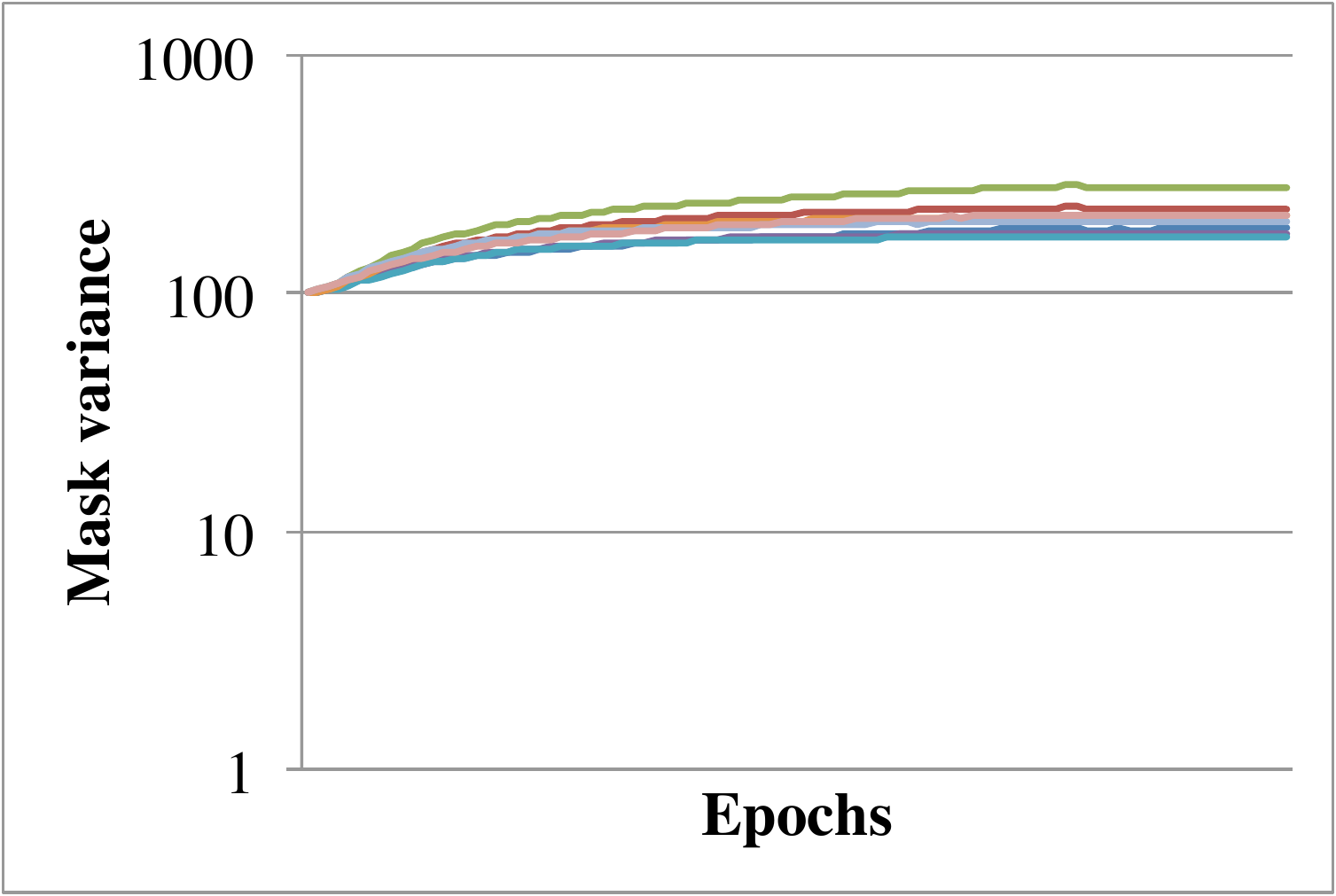}
  \caption{Evolution of the variance parameters for each of the 8 attention heads over course of training (left: first layer, right: second layer).
  }
  \label{fig:var}
\end{figure}

The Gaussian mask allows inspecting its trainable variance parameter. Fig.~\ref{fig:var} shows how the parameter evolves when initialized to a large value. It can be seen that in the first layer, diversity seems to be desirable, with some attention heads focusing on a small local context, and others on larger contexts. In contrast, the second layer does not appear to benefit from limiting its context. This partly confirms the idea of hierarchical modeling, where the modeling granularity increases across layers, but also shows that even at the bottom layer a controlled amount of long-range context is desirable.

\begin{table}[t]
  \centering
  \caption{WER results on attention biasing.}
\begin{tabular}{|c|c|c|c|c|c|}
\hline 
model & dev & test \tabularnewline 
\hline 
\hline 
stacked hybrid & 16.38 & 17.48  \tabularnewline 
\hline 
+ local masking & 15.42 & 16.17  \tabularnewline 
\hline 
+ Gauss mask (init.\ small) & 16.05 & 16.96 \tabularnewline  
\hline 
+ Gauss mask (init.\ large) & 14.90 & 15.89  \tabularnewline  
\hline 
\hline 
interleaved hybrid & 15.29 & 16.71  \tabularnewline 
\hline 
+ local masking & 15.44 & 16.19  \tabularnewline 
\hline 
+ Gauss mask (init.\ small) & 16.43 & 16.89  \tabularnewline 
\hline 
+ Gauss mask (init.\ large) & 15.00 & 15.82  \tabularnewline 
\hline 
\end{tabular}
\label{tab:biased_att}
\end{table}

\section{Interpretability of Attention Heads}

We hypothesize that certain attention heads respond to certain types of acoustic events. To test this hypothesis, we correlate the average attention that each attention head places on frames with the corresponding phoneme labels obtained via forced decoding. We re-train the stacked hybrid model with phonemes instead of characters as targets, and use encoder-decoder attention scores, summed over phoneme types, to obtain a soft alignment of phoneme labels for each frame. 
This gives us a measure for how much each frame in the sequence corresponds to the phoneme type under inspection.

We now correlate these phoneme activations to each of the first layer's 8 attention heads. We average the matrices across rows to obtain the overall attention that each frame receives. We then compute the Pearson correlation coefficient of the summarized self-attention and encoder-decoder attention sequences, concatenated over utterances.

Table~\ref{tab:analysis} shows the most highly correlated phonemes for each attention head, along with an attempt to classify these manually according to linguistic categories. This works remarkably well and we can clearly see a linguistically plausible division of labor, even though categories are neither exhaustive nor disjunct. Notice that head 2 seems to always focus on the utterance end where we usually expect silence, and head 8 is mostly unfocused, which we may interpret as these heads establishing channel and speaker context.


\begin{table}[t]
  \centering
  \caption{Analysis of function of attention heads. Note that we conducted a small amount of cherry picking by removing 4 outliers that did not seem to fit categories (OY from head 1, ZH from head 3, EH and ER from head 7). Entropy is computed over the correlation scores, truncated below 0.}
\begin{tabular}{|c|c|c|c|c|}
\hline 
$i$ & top phonemes & entropy & comments \tabularnewline
\hline 
\hline 
1 & S, TH, Z               &   3.7      & sibilants                                       \tabularnewline 
\hline 
2 & $<$/s$>$              &  1.9     & silence                                         \tabularnewline
\hline 
3 & UW, Y, IY, IX         &  3.6     & "you" diphthong \tabularnewline 
\hline 
   & B, G, D                 &        & voiced plosives \tabularnewline
\hline 
   & M, NG, N              &        & nasals \tabularnewline
\hline 
4 & XM, AW, AA, AY,   & 3.2  & A, schwa \tabularnewline
   & L, AO, AH & &  \tabularnewline
\hline 
5 & ZH, AXR, R          &  3.5       & R, ZH \tabularnewline
\hline 
6 & ZH, Z, S               &  3.2       & sibilants \tabularnewline
\hline 
   & IY, IH, Y, UW        &         & "you" diphthong \tabularnewline
\hline 
7 & S, $<$/s$>$, TH   & 3.4   & fricative, noise \tabularnewline 
   & CH, SH, F & &  \tabularnewline
\hline 
8 & mixed         & 3.7   &  unfocused \tabularnewline
\hline 
\end{tabular}
\label{tab:analysis}
\end{table}

\section{Conclusion}

Applying self-attention to acoustic modeling is challenging for computational and modeling reasons. We investigate ways to address these challenges and obtain our best results when using a hybrid model architecture and Gaussian biases that allow controlling context range. This model is almost as good as a strong LSTM-based baseline at much faster computation speed. We highlight interpretability as an advantage over conventional models. Future work includes investigation of self-attention with other sequences of low-information states such as characters, and of transferring results on controlling context range and interpretability to text modeling.

\bibliographystyle{IEEEtran}

\bibliography{library}

\end{document}